\documentclass[10pt,twocolumn,letterpaper]{article}

\usepackage{graphicx} 
\usepackage{multirow}
\usepackage[accsupp]{axessibility}

\usepackage{cvpr}              

\definecolor{cvprblue}{rgb}{0.21,0.49,0.74}
\usepackage[pagebackref,breaklinks,colorlinks,allcolors=cvprblue]{hyperref}

\def\paperID{*****} 
\def\confName{CVPR}
\def\confYear{2025}

\title{Heterogeneous Skeleton-Based Action Representation Learning}

\author{Hongsong Wang$^{1,2}$ \quad Xiaoyan Ma$^{3}$ \quad Jidong Kuang$^{3}$ \quad Jie Gui$^{3,4,5}$ \thanks{Corresponding Author} \\
	$^{1}$School of Computer Science and Engineering, Southeast University, Nanjing 210096, China \\
	$^2$Key Laboratory of New Generation Artificial Intelligence Technology and Its Interdisciplinary \\
	Applications (Southeast University), Ministry of Education, China \\ 
	$^{3}$School of Cyber Science and Engineering, Southeast University, Nanjing 210096, China\\
	$^{4}$Engineering Research Center of Blockchain Application, Supervision And Management\\ (Southeast University), Ministry of Education, China \\
	$^{5}$Purple Mountain Laboratories, Nanjing 210000, China \\
	\tt\small\{hongsongwang, 220224977, jidongkuang, guijie\}@seu.edu.cn \\
}

\begin{document}
\maketitle

\begin{abstract}
	Skeleton-based human action recognition has received widespread attention in recent years due to its diverse range of application scenarios. Due to the different sources of human skeletons, skeleton data naturally exhibit heterogeneity. The previous works, however, overlook the heterogeneity of human skeletons and solely construct models tailored for homogeneous skeletons. This work addresses the challenge of heterogeneous skeleton-based action representation learning, specifically focusing on processing skeleton data that varies in joint dimensions and topological structures. The proposed framework comprises two primary components: heterogeneous skeleton processing and unified representation learning. The former first converts two-dimensional skeleton data into three-dimensional skeleton via an auxiliary network, and then constructs a prompted unified skeleton using skeleton-specific prompts. We also design an additional modality named semantic motion encoding to harness the semantic information within skeletons. The latter module learns a unified action representation using a shared backbone network that processes different heterogeneous skeletons. Extensive experiments on the NTU-60, NTU-120, and PKU-MMD \uppercase\expandafter{\romannumeral2} datasets demonstrate the effectiveness of our method in various tasks of action understanding. Our approach can be applied to action recognition in robots with different humanoid structures.
\end{abstract}

\section{Introduction}
\label{sec:intro}

\begin{figure}[tp]
	\centering
	\includegraphics[width=0.9\linewidth]{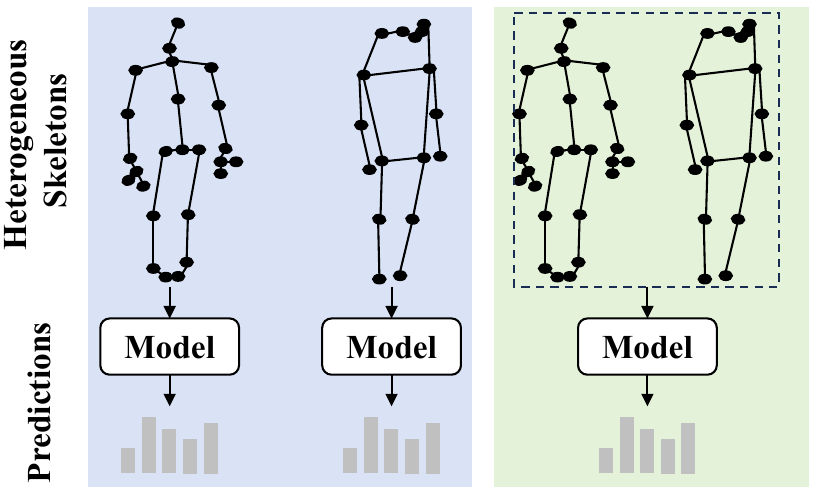}
	\caption{Comparison between our approach (right) and existing works (left). Previous works train specific models for each type of heterogeneous skeletons, whereas we design a unified model for heterogeneous skeletons.}
	\label{fig:intro}
	\vspace{-0.2cm}
\end{figure}

With the rapid advancement of sensors and detection algorithms, acquiring accurate human poses has become easier. Skeleton-based action recognition has emerged as a prevalent task in the field of computer vision. It holds a significant role in areas such as human-computer interaction, medical rehabilitation, video surveillance and intelligent sports.

Skeletons are a commonly used modality in action recognition tasks. Compared to other modalities, such as videos and depth map sequences, skeletons possess characteristics of high abstraction, low complexity, and good robustness. Furthermore, skeletons naturally align with human actions in a physical sense, allowing for a better representation of human motion. Skeletons can be collected by numerous sensors, thus resulting in heterogeneity in skeleton data. The heterogeneity of skeleton data is mainly manifested in the differences in coordinate dimensions and the number of joints defining the human body structure. For example, the skeleton data recorded by the Kinect V2 depth sensor is a three-dimensional sequence featuring 25 joints, whereas the skeleton data estimated from the RGB video is typically a two-dimensional sequence with 17 joints. Addressing data heterogeneity in human action recognition will emerge as a significant research topic.

Although deep learning methods have advanced rapidly in the field of skeleton-based human action recognition, with many excellent approaches emerging, these methods primarily focus on proposing different network architectures for single homogeneous data. For instance, these methods can be broadly classified into three categories: Recurrent Neural Networks (RNNs)-based~\cite{du2016representation,wang2017modeling}, Graph Convolutional Networks (GCNs)-based~\cite{yan2018spatial,xiang2023generative,zhou2024blockgcn}, and Transformers-based~\cite{zhu2023motionbert,zhou2022hypergraph}. Most of these methods rely on supervised training, requiring the model to be trained from scratch for each specific task. Consequently, these models lack sufficient transferability across different datasets. Recently, self-supervised skeleton-based action recognition~\cite{guo2022contrastive, zhang2022contrastive, sun2023unified} has attracted considerable attention owing to its enhanced transferability across various datasets. However, these methods overlook the heterogeneity of data, therefore, the model can only demonstrate transferability across datasets that contain homogeneous skeleton data.

This paper addresses the heterogeneity of human skeleton data, and our goal is to design a unified human action recognition model for heterogeneous data. The diagram of the proposed learning paradigm is illustrated in Figure~\ref{fig:intro}. Compared with existing works that train individual models for different types of heterogeneous data, our method is capable of training a unified model encompassing these data. Our method primarily comprises heterogeneous skeleton processing and unified representation learning. The former converts heterogeneous data into a unified format, while the latter learns a unified action representation for different types of skeleton data. 

More specifically, we focus on the two most commonly used types of data: the three-dimensional 25-joint skeleton and the two-dimensional 17-joint skeleton. To unify dimensions, we design a 3D pose estimation module to convert the two-dimensional skeleton into a three-dimensional skeleton, as the three-dimensional data contains richer information about human action. To unify the skeleton topology, we construct a prompted unified skeleton by selecting common joints from different heterogeneous skeletons and designing trainable skeleton-specific prompts to complement the missing joints for each type of skeleton. Since skeletons of different structures all semantically represent the human body but lack semantic information in their coordinate joints, we design semantic motion encoding, which utilizes pretrained language models to encode the semantic information to aid in action representation learning. The unified representation learning comprises an efficient Transformer-based architecture designed to learn unified representations from diverse types of data.

In summary, our main contributions are as follows:
\begin{itemize}
	\item To the best of our knowledge, this is the first work that studies data heterogeneity in human skeletons and proposes a unified framework for learning heterogeneous skeleton-based action representations.
	\item We propose a heterogeneous skeleton processing module to unify heterogeneous skeletons from both the perspectives of coordinate dimensions and skeleton topologies. 
\end{itemize}

\section{Related Work}
\label{sec:related}

\begin{figure*}[tp]
	\centering
	\includegraphics[width=\linewidth]{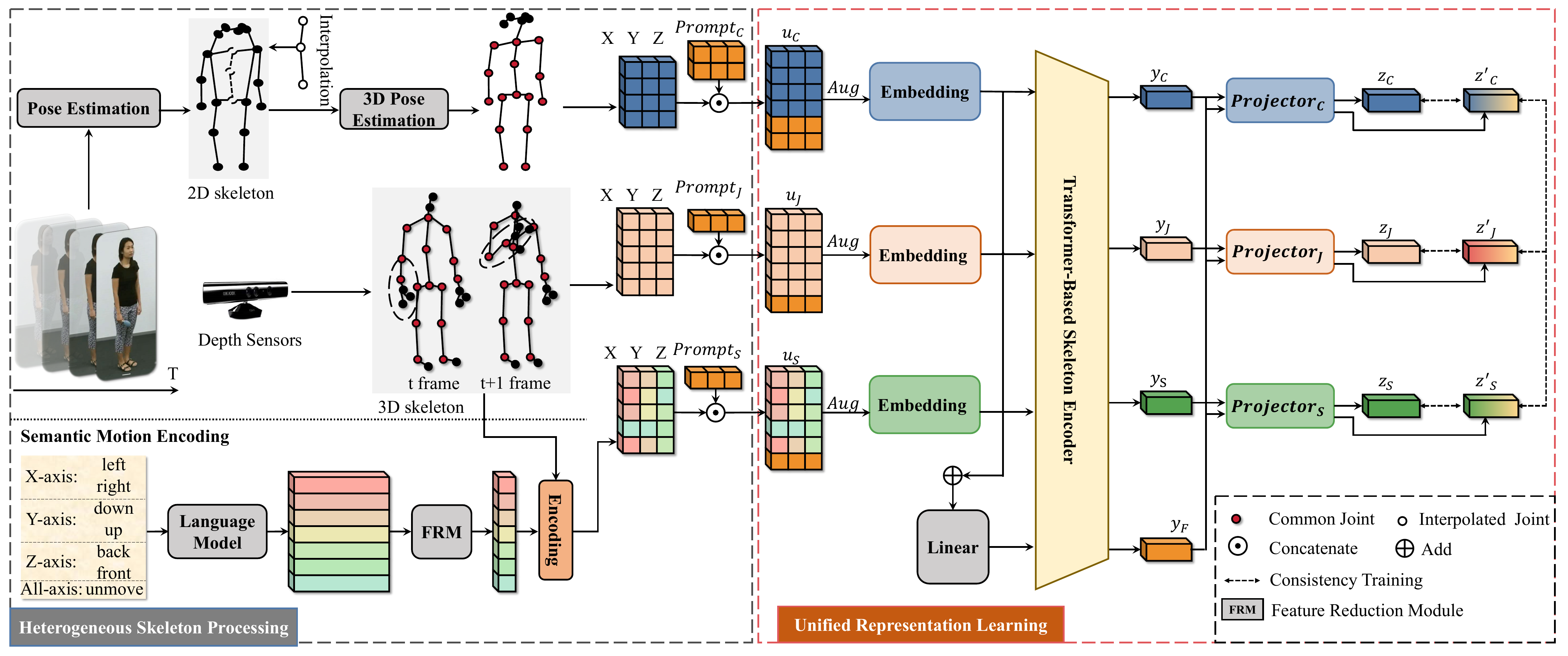}
	\caption{The structure of our framework. This framework comprises heterogeneous skeleton processing and unified representation learning. Initially, we convert the two-dimensional skeleton extracted by the pose estimation model into three dimensions using a 3D pose estimation module. Subsequently, we employ trainable skeleton-specific prompts to construct a prompted, unified skeleton representation. Additionally, we design semantic motion encoding to enrich the representation learning process with more semantic information. The unified representation learning trains a unified action representation network for these processed heterogeneous skeletons.}
	\label{fig:framework}
\end{figure*}

\noindent\textbf{Self-Supervised Skeleton-based Action Recognition: }
Self-supervised skeleton-based action recognition~\cite{zheng2018unsupervised,su2020predict,nie2020unsupervised,xu2021unsupervised} refers to the use of unlabeled skeleton to perform action recognition tasks and has become a research hotspot in recent years. 
Su et al.~\cite{su2020predict} train an encoder-decoder network in an unsupervised manner to relate skeleton sequences with actions. Nie et al.~\cite{nie2020unsupervised} propose a Siamese denoising autoencoder for 3D human pose representation learning. Xu et al.~\cite{xu2021unsupervised} introduce a Motion Capsule Autoencoder (MCAE) to address transformation invariance in unsupervised learning. In recent years, more and more studies adopt contrastive learning methods for unsupervised learning~\cite{su2021self,wang2022contrast,chi2022infogcn}. Contrastive methods apply various augmentations to the unlabeled skeleton sequences, generating views that form positive and negative sample pairs. These pairs are then used to train the model to pull positive pairs closer and push negative pairs apart. 
Su et al.~\cite{su2021self} construct fragments with speed variations and motion interruptions to determine positive and negative samples for contrastive learning. Wang et al.~\cite{wang2022contrast} perform contrastive learning on representations learned from both skeleton coordinate sequences and velocity sequences. 
Additionally, some studies introduce other techniques to enhance model performance~\cite{chen2022hierarchically,dong2023hierarchical,lin2023actionlet,wu2024scd,shah2023halp,mao2023masked,zhu2023modeling}. Chen et al.~\cite{chen2022hierarchically} employ a hierarchical pre-training approach to enhance action representation capabilities. Dong et al.~\cite{dong2023hierarchical} generate multiple features at different granularities to perform contrastive learning in a hierarchical manner. Lin et al.~\cite{lin2023actionlet} transform the data into actionlet and non-actionlet regions to enhance contrastive learning ability. Wu et al.~\cite{wu2024scd} extend contrastive loss to measure spatiotemporal representation differences and use a masking strategy to increase the diversity of training data. Weng et al.~\cite{weng2024usdrl} present a unified dense representation learning framework based on feature decorrelation. Despite the significant progress made by the above methods in advancing self-supervised skeleton action recognition, most of the research focuses on uni-modal data, failing to effectively explore the complementarity among multimodal skeleton data.

\noindent\textbf{Multi-modal Action Representation Learning: }
Multimodal information of skeletons, such as joints and bones, has been verified to exhibit strong complementarity for human action understanding~\cite{wang2018beyond}.
To improve the representation capability of actions, recent works utilize multi-modal data to encourage models to learn cross-modal features. For example, Xiang et al.~\cite{xiang2023generative} propose a Generative Action Prompt (GAP) method, which leverages action semantic information to enhance the representation learning ability of the skeleton encoder. Chi et al.~\cite{chi2022infogcn} incorporate the relative positions of joints to augment the multi-modal representation of the skeleton, providing complementary spatial information for the joints. Sun et al.~\cite{kim2022global} employ an early fusion strategy to input joint, motion, and bone modalities into the same stream, thereby reducing model complexity. In addition, a common approach for utilizing multi-modal data is to extend uni-modal methods to multi-modal ones through late fusion strategies~\cite{zhang2023hierarchical,guo2022contrastive,zhou2023self}. This approach independently trains multiple single-modal models and then fuses their outputs to enhance representational capacity~\cite{li20213d,zhang2022contrastive,mao2022cmd}. While these methods adopt multi-modal skeleton inputs, they fail to effectively handle and exploit heterogeneous data with complementary information, such as fine-grained semantics and skeletons with different topologies. Different from these approaches, we propose is a multi-modal representation learning framework that processes heterogeneous skeletons with different dimensions, topologies, and modalities through a unified network. 

\section{Method}
\label{sec:method}

\subsection{Heterogeneous Skeleton Processing}
\begin{figure}[t]
	\centering
	\includegraphics[width=\linewidth]{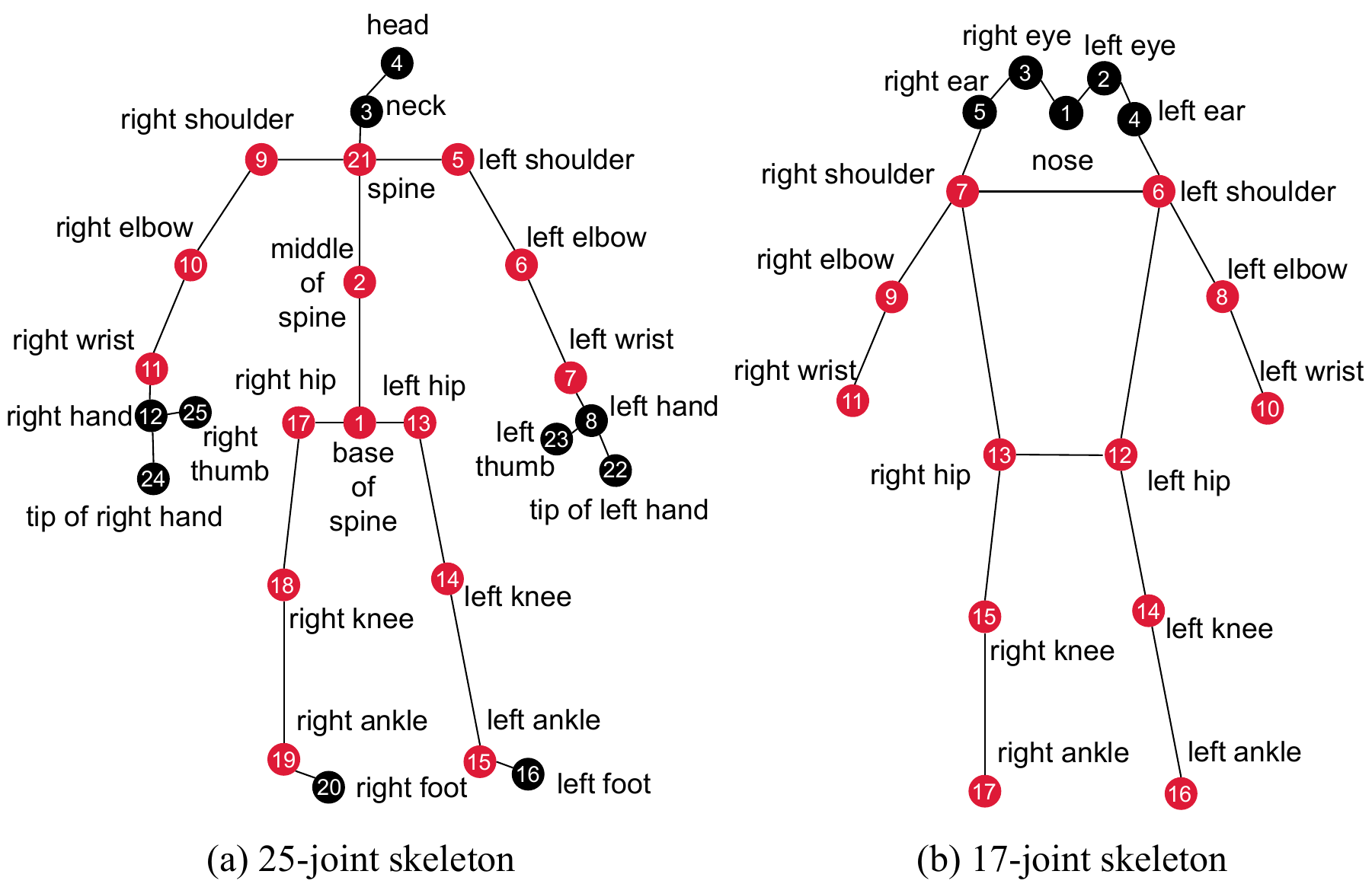}
	\caption{Comparison between two popular human skeletons: the 25-joint skeleton and the 17-joint skeleton. The name of each human joint is annotated near the index. Joints that are common to both skeletons are marked in red, while unique joints are marked in black. We consider three joints in the spine of the 25-joint skeleton as common joints because they can be easily interpolated by surrounding joints in the 17-joint skeleton.}
	\label{fig:skeleton}
\end{figure}

Human skeleton data from different sources exhibit heterogeneity in skeleton structure. This heterogeneity is primarily manifested in two aspects: varying numbers of human body nodes and differing coordinate dimensions. For example, the human skeleton data collected by the Kinect V2 depth sensor consists of 3D data for 25 joints, whereas the human skeleton estimated from RGB video typically consists of 2D data for 17 joints. Each joint represents a distinct part of the human body. The comparison between the 25-joint skeleton and the 17-joint skeleton is in Figure~\ref{fig:skeleton}. The 25-joint skeleton includes more joints in the hands, whereas the 17-joint skeleton has more joints focused on the face. We use these two skeletons as illustrative examples to process heterogeneous skeletons. 

\noindent\textbf{3D Pose Estimation:} 
Three-dimensional skeletons contain more skeleton information compared to two-dimensional skeletons, which can better assist models in extracting features. Therefore, we aim to reconstruct the two-dimensional 17-joint skeleton into a three-dimensional one.

Since the spine lies on the central axis of the body, the missing three spine joints in the 17-joint skeleton can be interpolated using the joints of the left and right shoulders and hips. Thus, we manually add 3 spine-related joints to the 17-joint skeleton using linear interpolation to facilitate 3D pose estimation. The preprocessing step for interpolation is as follows:
\begin{align}
	&p_{spine} = (p_{left\_shoulder} + p_{right\_shoulder})/2, \\
	&p_{base\_of\_spine} = (p_{left\_hip} + p_{right\_hip})/2, \\
	&p_{middle\_of\_spine} = (p_{spine} + p_{base\_of\_spine})/2,
\end{align}
where $p$ represents the coordinates of a particular joint. After adding spine-related joints, the original 17-joint skeleton now has a total of 20 joints, but for the sake of continuity and simplicity, we still refer to it as the 17-joint skeleton.

After joint interpolation, a 3D pose estimation module is designed to predict the three-dimensional skeleton given the two-dimensional skeleton. By interpolating additional spine-related joints, more paired two-dimensional and three-dimensional joints are also made available during the training of the 3D pose estimation module.

\noindent\textbf{Prompted Unified Skeleton:} As shown in Figure \ref{fig:skeleton}, these two skeletons have different numbers of joints. So we has a total of 30 different joints. In order to process skeletons with differing numbers of joints, a unified skeleton format is necessary. Therefore, we use the method of prompt learning to define two trainable skeleton-specific prompts corresponding to the 25-joint skeleton and 17-joint skeleton: \(prompt_{J}\in \mathbb{R}^{5\times 3}\) and \(prompt_{C}\in \mathbb{R}^{10\times 3}\), which are used to construct prompted unified skeleton \(u\in \mathbb{R}^{m\times t\times 30\times 3}\). 

The prompt and the skeleton are concatenated in the dimension of the joint. Moreover, the original joint order differs between the 25-joint skeleton and the 17-joint skeleton. Therefore, we adjust them to a unified order, which is divided into three parts: human facial joints numbered from 1 to 5, common joint points numbered from 6 to 20, and head-hand-foot joint points numbered from 21 to 30.

\subsection{Semantic Motion Encoding}
The coordinates of the joint only represent physical motion of human action. To learn a unified model for heterogeneous skeletons, it is also crucial to harness the semantic information of the body within these skeletons. However, the semantics of body joints constitute static information regarding the names of the joints, which does not reflect the action being performed. To address this issue, we encode the semantics of joint motion as auxiliary information to assist in action representation learning.

Changes in the direction of motion during actions reflect the trend of motion and contain rich dynamic information. Therefore, they can serve as dynamic information to help process of Semantic Motion. The motion of joints can be divided into three directions: x-axis, y-axis, and z-axis. We use ``right" and ``left" to represent the direction on the x-axis; use ``up" and ``down" to represent the direction on the y-axis; use ``front" and ``back" represent the direction on the z-axis; finally, use ``unmove" to represent no motion, with a total of seven directions of motion. 

These seven direction word are fed into a pre-trained language model to get high-dimensional semantic features $e \in \mathbb{R}^{7\times l}$, where $l$ represents the length of the feature vector, which depends on the specific language model used. The high-dimensional semantic features make it difficult for the model to learn effective action representation, and the high dimensionality can cause excessive computational pressure. A feature reduction module is designed to map these features to a low-dimensional encoding. We set the embedding to 1 to reduce dimensionality, so that we can construct semantic motion encoding with the same size as the three-dimensional skeleton.

The semantic motion encoding is obtained by:
\begin{align}
	m^x_{{t,j}}=\left\{
	\begin{array}{ll}
		\tilde{e}_{left}&s^x_{t,j} - s^x_{t-1,j} <0,\\ 
		\tilde{e}_{ummove}&s^x_{t,j} - s^x_{t-1,j} =0,\\
		\tilde{e}_{right}&s^x_{t,j} - s^x_{t-1,j} >0,
	\end{array}
	\right. \\
	m^y_{t,j}=\left\{
	\begin{array}{ll}
		\tilde{e}_{down}&s^y_{t,j} - s^y_{t-1,j}<0,\\ 
		\tilde{e}_{ummove}&s^y_{t,j} - s^y_{t-1,j}=0,\\
		\tilde{e}_{up}&s^y_{t,j} - s^y_{t-1,j}>0,
	\end{array}
	\right. \\
	m^z_{t,j}=\left\{
	\begin{array}{ll}
		\tilde{e}_{back}&s^z_{t,j} - s^z_{t-1,j}<0,\\ 
		\tilde{e}_{ummove}&s^z_{t,j} - s^z_{t-1,j}=0,\\
		\tilde{e}_{front}&s^z_{t,j} - s^z_{t-1,j}>0,
	\end{array}
	\right.
\end{align}
where $\tilde{e}$ is the semantic feature after dimensionality reduction, $s^x_{t,j}$ and $m^x_{t,j}$ represent the original value and the semantic motion encoding of the x-coordinate of the $j$-th body joint in the $t$-th frame, respectively.

Similar to heterogeneous skeleton processing, semantic motion encoding is also converted into a unified format utilizing prompt embeddings. This operation is as follows:
\begin{equation}
	u_{S}=prompt_{S}\odot s_{S}
\end{equation}
where $\odot$ denotes the concatenation operation, $s_{S}$ and $u_{S}$ represent the semantic motion encoding before and after this operation, respectively. As this prompted encoding has the same size as the real skeletons, it can perform the same data augmentation operations. 

\subsection{Unified Representation Learning}
We design a unified network to learn action representations from processed heterogeneous skeletons. We adopt the self-supervised learning paradigm due to its capabilities of transferability and generalization. We use the Transformer-based architecture~\cite{plizzari2021skeleton} as the feature backbone.

A specific embedding layer is first used to map each type of skeleton to the embeddings. Then, we follow the early fusion strategy~\cite{sun2023unified} to train the network with inputs of heterogeneous skeletons. Let \(h_{J}\), \(h_{C}\), and \(h_{S}\) into \(h_{F}\) be the embeddings of the 25-joint skeleton, 17-joint skeleton and semantic motion. The fused embedding $h_{F}$ is computed as:
\begin{equation}
	h_{F}=linear(\frac{1}{3}(h_{J}+h_{C}+h_{S})),
\end{equation}
where \(linear(\cdot)\) is learnable linear transformation. 


After early fusion of embeddings, a unified feature encoder is used to produce skeleton feature \(y\in \mathbb{R}^{D}\), where $D$ denotes the feature dimension. This process is:
\begin{equation}
	y=encoder(h),
\end{equation}
where $h \in \{h_{J}, h_{C}, h_{S}, h_{F} \}$, and $y$ is the corresponding feature of $h$. 

\noindent\textbf{Feature Consistency Loss:} 
The unsupervised training losses of feature consistency include two levels: (1) between \(y_{F}\) and \(\{y_{J},y_{C},y_{S}\}\); (2) among the elements of the set \(\{y_{J},y_{C},y_{S}\}\). During training, given a batch of features, \(Y\in \mathbb{R}^{N\times 4\times D}\), \(Y\) is mapped into skeleton-specific spaces by skeleton-specific projectors. Consequently, the consistency learning loss is defined as:
\begin{equation}
	\begin{split}
		\mathcal{L}_{con}=\sum_{i}MSE(Z_{i},Z'_{i}) +\sum_{i\ne j} MSE(Z_{i},Z_{j}),
	\end{split}
\end{equation}
where \(i,j\in\{J,C,S\}\), \(Z\) and \(Z'\) are skeleton-specific representations obtained from specific skeleton feature and fused feature, respectively. 

\noindent\textbf{3D Pose Estimation Loss:} In heterogeneous skeleton processing, we design a 3D pose estimation module to regress the three-dimensional skeleton from the two-dimensional one. This module is jointly trained with the unified action representation learning. The loss for 3D pose estimation is formulated as: 
\begin{equation}
	\begin{split}
		\mathcal{L}_{rec} =\frac{1}{\vert \mathcal{B} \vert} \sum_{i \in \mathcal{B}} \left \|u^C_i-u^J_i \right \|^{2}_{2},
	\end{split}
\end{equation}
where $\mathcal{B}$ is the set of common joints between the 25-joint skeleton and the 17-joint skeleton, $u^C$ and $u^J$ denote the corresponding prompted unified skeletons for the 17-joint and 25-joint skeletons, respectively.

\noindent\textbf{Regularizations:} 
We use VICREG ~\cite{bardes2021vicreg} for action representation learning. VICREG includes semantic consistency regularization and variance-covariance (VC) regularization. 
The variance-covariance (VC) regularization include a variance term and a covariance term. The variance term is:
\begin{equation}
	V(Z)=\frac{1}{D}\sum_{j=1}^{D}max(0,\gamma -\sqrt{Var(Z_{:,j})} +\epsilon),
\end{equation}
where \(\gamma\) is variance threshold, \(\epsilon\) is a small scalar preventing numerical instabilities, and \(Var(Z_{:,j})\) is the variance of jth embedding dimension vector \(Z_{:,j}\). The covariance term is:
\begin{equation}
	C(Z)=\frac{1}{D}\sum_{i\ne j}[Cov(Z)]^{2}_{i,j},
\end{equation}
where \(Cov(Z)\) is the auto-covariance matrix of \(Z\). The final VC regularization loss is:
\begin{equation}
	\mathcal{L}_{reg}=\sum_{i}VC(Z_{i})+VC(Z'_{i}),
\end{equation}
where \(i\in\{J,C,S\}\), and $VC(Z)$ is formulated as:
\begin{equation}
	VC(Z)=\mu V(Z)+C(Z).
\end{equation}

The total training objective loss is:
\begin{equation}
	\mathcal{L}=\lambda \mathcal{L}_{con} + \mathcal{L}_{reg} + \mathcal{L}_{rec}.
\end{equation}

\begin{table*}[t]
	\centering
	\resizebox{0.95\textwidth}{!}{
	\begin{tabular}{lcclccccc} 
		\toprule
		\multirow{2}{*}{\textbf{Method}} & \multirow{2}{*}{\textbf{Publication}} & \multirow{2}{*}{\textbf{Modality}} & \multirow{2}{*}{\textbf{FLOPs/G}} & \multicolumn{2}{c}{\textbf{NTU-60}} & \multicolumn{2}{c}{\textbf{NTU-120}} & \textbf{PKU-MMD II}  \\ 
		\cmidrule(l){5-8}\cmidrule(lr){9-9}
		&                                       &                                    &                                   & x-sub         & x-view              & x-sub         & x-set                & x-sub                \\ 
		\midrule
		AimCLR~\cite{guo2022contrastive}      & AAAI'22                               & J                                  & 1.15                              & 74.3          & 79.7                & 63.4          & 63.4                 & --                   \\
		PTSL~~\cite{zhou2023self}             & AAAI'23                               & J                                  & \textbf{1.15}                     & 77.3          & 81.8                & 66.2          & 67.7                 & 49.3                 \\
		GL-Transformer~\cite{kim2022global}   & ECCV'22                               & J                                  & 118.62                            & 76.3          & 83.8                & 66.0          & 68.7                 & --                   \\
		CPM~\cite{zhang2022contrastive}       & ECCV'22                               & J                                  & 2.22                              & 78.7          & 84.9                & 68.7          & 69.6                 & 48.3                 \\
		CMD~\cite{mao2022cmd}                 & ECCV'22                               & J                                  & 5.76                              & 79.8          & 86.9                & 70.3          & 71.5                 & 43.0                 \\
		IGM~\cite{lin2025idempotent}                 & ECCV'24                               & J                                  & --                              & 86.2          & 91.2                & 80.0          & 81.4                 & --                 \\
		HYSP~\cite{francohyperbolic} & ICLR'23  &  J & -- & 78.2  & 82.6 & 61.8  & 64.6 & -- \\
		UmURL~\cite{sun2023unified}           & ACM MM'23                             & J                                  & 1.74                              & 82.3          & 89.8                & 73.5          & 74.3                 & 52.1                 \\
		PCM\(^3\)~\cite{zhang2023prompted}           & ACM MM'23                             & J                                  & --                              & 83.9          & 90.4                & 76.5          & 77.5                 & 51.5                 \\
		RVTCLR+~\cite{zhu2023modeling}           & ICCV'23                             & J                                  & --                              & 74.7          & 79.1                & ---          & --                 & --                 \\
		HaLP~\cite{shah2023halp}           & CVPR'23                             & J                                  & --                              & 79.7          & 86.8                & 71.1          & 72.2                 & 43.5                 \\
		ActCLR~\cite{lin2023actionlet}           & CVPR'23                             & J                                  & --                              & 80.9          & 86.7                & 69.0          & 70.5                 & --                 \\
		USDRL~\cite{weng2024usdrl} & AAAI'25 & J & -- & 84.2 & 90.8 & 76.0 & 76.9 &  51.8 \\
		\midrule
		3s-AimCLR~\cite{guo2022contrastive}   & AAAI'22                               & J\,+\,M\,+\,B                      & 3.45                              & 78.9          & 83.8                & 68.2          & 68.8                 & 39.5                 \\
		3s-CPM~\cite{zhang2022contrastive}    & ECCV'22                               & J\,+\,M\,+\,B                      & 6.66                              & 83.2          & 87.0                & 73.0          & 74.0                 & 51.5                 \\
		3s-CMD~\cite{mao2022cmd}              & ECCV'22                               & J\,+\,M\,+\,B                      & 17.28                             & 84.1          & 90.9                & 74.7          & 76.1                 & 52.6                 \\
		3s-HiCLR~\cite{zhang2023hierarchical} & AAAI'23                               & J\,+\,M\,+\,B                      & 7.08                              & 78.8          & 83.1                & 67.3          & 69.9                 & -                    \\
		3s-PSTL~\cite{zhou2023self}           & AAAI'23                               & J\,+\,M\,+\,B                      & 3.45                              & 79.1          & 83.8                & 69.2          & 70.3                 & 52.3                 \\
		3s-HYSP~\cite{francohyperbolic} & ICLR'23  &  J\,+\,M\,+\,B & -- & 79.1 & 85.2 & 64.5 & 67.3 & -- \\
		UmURL~\cite{sun2023unified}           & ACM MM'23                             & J\,+\,M\,+\,B                      & \textbf{2.54}                     & 84.2          & 90.9                & 75.2          & 76.3                 & 54.0                 \\
		3s-UmURL~\cite{sun2023unified}        & ACM MM'23                             & J\,+\,M\,+\,B                      & 5.22                              & 84.4          & 91.4                & 75.9          & 77.2                 & 54.3                 \\
		3s-RVTCLR+~\cite{zhu2023modeling}           & ICCV'23                         &J\,+\,M\,+\,B                                  &--                               & 79.7          & 84.6                & 68.0          & 68.9                 & --                \\
		3s-ActCLR~\cite{lin2023actionlet}           & CVPR'23                             &J\,+\,M\,+\,B                                 & --                              & 84.3          & 88.8               & 74.3          & 75.7                 & --                 \\
		USDRL~\cite{weng2024usdrl} & AAAI'25 & J\,+\,M\,+\,B & --  & 87.1 & 93.2 & 79.3 & 80.6 & 59.7 \\
		\midrule
		Ours                                  & –                                     & J                                  & 1.74                              & 80.2          & 88.0                & 70.7          & 73.5                 & 47.7                 \\
		Ours                                  & –                                     & C                                  & \textbf{1.74}                              & \textbf{84.4}          & \textbf{90.6}                &  \textbf{73.5}          & \textbf{78.4}                 & \textbf{54.1}                 \\
		Ours                                  & –                                     & S                                  & 1.74                              & 70.1          & 75.7                & 58.3          & 60.2                 & 33.8                 \\ 
		Ours                                  & --                                    & J\,+\,C                            & 2.17                              & 86.1          & 92.7                & 75.8          & 80.0                 & 57.3                 \\
		Ours                                  & --                                    & J\,+\,S                            & 2.17                              & 80.7          & 88.0                & 71.0          & 73.2                 & 48.9                 \\
		Ours                                  & --                                    & C\,+\,S                            & \textbf{2.17}                     & 85.0          & 90.3                & 73.8          & 78.3                 & 54.1                 \\
		Ours                                  & --                                    & J\,+\,C\,+\,S                      & \textbf{2.54}                     & \textbf{87.8} & \textbf{93.7}       & \textbf{78.9} & \textbf{82.2}        & \textbf{58.2}        \\
		\bottomrule
	\end{tabular}
    }
	\caption{Comparison with state-of-the-art methods on the skeleton-based action recognition task. Uni-modality and multi-modal methods are compared on the NTU-60, NTU-120, and PKU-MMD II datasets. For simplicity, J and C represent the joint modalities of a three-dimensional 25-joint skeleton and a two-dimensional 17-joint skeleton, respectively, while S signifies the semantic motion encoding.}
	\label{tab_recognition}
\end{table*}

\begin{table}[t]
	\centering
	\setlength\tabcolsep{3pt}
	\resizebox{0.48\textwidth}{!}{
	\begin{tabular}{lccccc} 
		\toprule
		\multirow{2}{*}{\textbf{Method}} & \multirow{2}{*}{\textbf{Modality}} & \multicolumn{2}{c}{\textbf{NTU-60}}    & \multicolumn{2}{c}{\textbf{NTU-120}}    \\ 
		\cmidrule(lr){3-4}\cmidrule(lr){5-6}
		&                           & x-sub         & x-view        & x-sub         & x-set        \\ 
		\midrule
		LongT GAN~\cite{zheng2018unsupervised}                & J                         & 39.1          & 48.1          & 31.5          & 35.5           \\
		P\&C~\cite{su2020predict}                             & J                         & 50.7          & 76.3          & 39.5          & 41.8           \\
		AimCLR~\cite{guo2022contrastive}                      & J                         & 62.0          & 71.5          & -             & -              \\
		ISC~\cite{thoker2021skeleton}                         & J                         & 62.5          & 82.6          & 50.6          & 52.3           \\
		HiCLR~\cite{zhang2023hierarchical}                    & J                         & 67.3          & 75.3          & -             & -              \\
		HiCo~\cite{dong2023hierarchical}                      & J                         & 68.3          & 84.8          & 56.6          & 59.1           \\
		CMD~\cite{mao2022cmd}                                 & J                         & 70.6          & 85.4          & 58.3          & 60.9           \\
		HaLP~\cite{shah2023halp}                              & J                        
		& 65.8          & 83.6          & 55.8          & 59.0           \\
		UmURL~\cite{sun2023unified}                           & J                         & 71.3          & 88.3          & 58.5          & 60.9           \\
		UmURL~\cite{sun2023unified}                           & J\,+\,M\,+\,B                     & 72.0          & 88.9          & 59.5          & 62.2          \\
		\midrule
		Ours                                                  & J                         & 66.3          & 87.1          & 55.7          & 59.8           \\
		Ours                                                  & C                         & 70.3          & 85.2          & 58.5          & 64.8           \\
		Ours                                                  & S                        & 53.5          & 74.8          & 43.4          & 46.2           \\ 
		Ours                                                  & J\,+\,C\,+\,S                    & \textbf{72.7} & \textbf{90.9} & \textbf{61.9} & \textbf{66.9} \\
		\bottomrule
	\end{tabular}}
	\caption{Comparison with state-of-the-art methods on skeleton-based action retrieval task on the NTU-60 and NTU-120 datasets.}
	\label{tab_retrieval}
\end{table}

\section{Experiments}
\label{sec:experiments}

\subsection{Experiments Settings}

\noindent\textbf{Dataset and Evaluation Metric:} 
NTU-60~\cite{liu2019ntu}, NTU-120~\cite{shahroudy2016ntu} and PKU-MMD II~\cite{liu2020benchmark} datasets are used to train and evaluate the model. Consistent with prior works, we use top-1 accuracy as the evaluation metric.


\noindent\textbf{Implementation Details:}
We employ HRNet~\cite{sun2019deep} for pose estimation to extract two-dimensional skeleton keypoints from the datasets. For semantic motion encoding, we utilize the pre-trained ViT-B/32 text encoder of the CLIP~\cite{radford2021learning}. In the 3D pose estimation module, we apply a 4-layer MLP to convert 2D skeleton coordinates to 3D, standardizing the input using BatchNorm and processing negative values with the LeakyReLU activation function. The encoder backbone incorporates two single-head Transformers to model the spatial and temporal dimensions independently, each with a hidden dimension of 1024. 
The entire experiment is conducted within the PyTorch framework and accelerated using two NVIDIA GeForce RTX 4090 GPUs.

\subsection{Comparison with The SOTA Methods}
After representation learning, our method is already capable of extracting expressive features. We compare our method with state-of-the-art methods on two tasks: skeleton-based action recognition and skeleton-based action retrieval.

\noindent\textbf{Skeleton-based Action Recognition:} We adopt the same practices as in previous works~\cite{li20213d,sun2023unified}. Specifically, we freeze the model weights and use them as an encoder. After that only a linear classifier is trained to classify the features extracted by the encoder. We conduct experiments by utilizing 3D Pose Estimation and Semantic Motion Encoding. Moreover, we calculate the computational complexity of the model. Experimental results are presented in Table~\ref{tab_recognition}.

The results show that our method significantly outperforms other approaches in heterogeneous skeleton applications, validating the effectiveness of the our method. Compared to the state-of-the-art 3s-UmURL~\cite{sun2023unified}, our model performs better when using the same number of skeleton. This advantage is primarily attributed to the finer-grained information of heterogeneous skeleton and richer dynamic semantic information, which allow the model to capture expressive features. In terms of computational efficiency, our model demonstrates lower FLOPs than CrossSCLR~\cite{li20213d} and CMD~\cite{mao2022cmd}, owing to the Unified Representation Learning. Our method also demonstrates excellent flexibility when dealing with a single skeleton or any combination of two sets of skeleton.

\noindent\textbf{Skeleton-based Action Retrieval:} In the skeleton-based action retrieval experiment, cosine similarity is employed for action query retrieval. Similarly, the pre-trained model is not fine-tuned. As shown in Table~\ref{tab_retrieval}, the experimental results demonstrate that our method achieves superior retrieval performance on the NTU-60 and NTU-120 datasets when processing heterogeneous skeletons. This further confirms that the proposed heterogeneous skeleton-based action representation learning significantly improves the performance of downstream tasks, clearly indicating that the model effectively extracts more discriminative and expressive features.

\noindent\textbf{Ablation Studies: }
The ablation experiments in Table~\ref{tab_ablation} include specifically removing the modules of 3D pose estimation, semantic motion encoding, and skeleton-specific prompt embedding during training. To study the impact of the 3D pose estimation module, we conduct a comparative analysis using only the original two-dimensional skeleton data. The results indicate that the reconstructed three-dimensional skeleton achieves superior performance. Additionally, to examine the effectiveness of semantic motion, we replace the semantic features extracted from a language model with a simple numeric representation of joint motion, where positive, negative, and no motion are represented by 1, -1, and 0, respectively. The results confirm that that semantic motion, enriched with detailed action-related information, improves the model’s recognition performance. Finally, to evaluate the role of skeleton-specific prompts, we replace traditional zero-padding with trainable prompts for comparison. The results show that trainable prompts enhance the representational capacity of the skeleton topology. Collectively, these enhancements lead to notable improvements in the model’s action recognition accuracy.

\begin{figure}
	\centering
	\includegraphics[width=\linewidth]{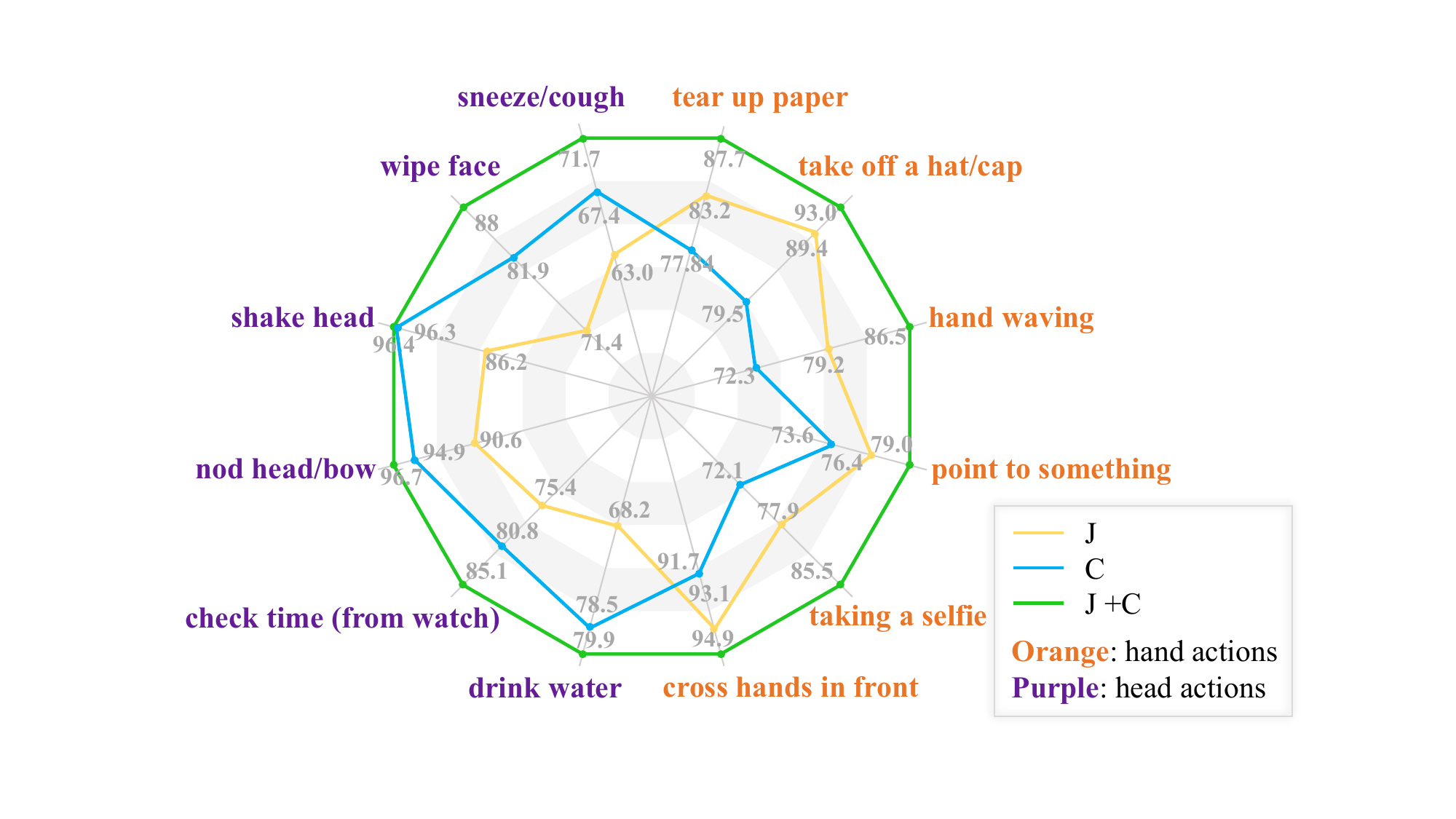}
	\caption{Comparison of skeletons in terms of single-class action accuracy. The experiment is conducted under the NTU-60/x-sub.}
	\label{fig:radar_chart}
\end{figure}

\noindent\textbf{Discussions:} We further discuss the roles of semantic motion encoding and heterogeneous skeleton in action recognition. On the one hand, as can be seen from Table~\ref{tab_recognition} and Figure \ref{fig:radar_chart}, semantic motion encoding generally improves recognition accuracy across all experimental settings, indicating that data from semantic modalities can effectively assist the model in recognizing actions. On the other hand, as previously analyzed, the 25-joint skeleton includes fine-grained hand and foot joints, which is advantageous for recognizing actions heavily influenced by these joints, such as hand-waving and pointing, as shown in the Figure \ref{fig:radar_chart}. In contrast, the 17-joint skeleton is better suited for recognizing actions dominated by facial joints, such as head-shaking and face-wiping. By integrating heterogeneous skeletons in a unified representation learning framework, we provide the model with a more comprehensive set of skeleton information, enabling it to capture richer and more detailed features for action recognition.

\begin{table}
	\centering
	\begin{tabular}{lc} 
		\toprule
		\textbf{Method} & \textbf{PKU-MMD II}  \\
		\midrule
		w/o 3D pose estimation                 & 55.8        \\
		w/o semantic motion                     & 57.9        \\
		w/o skeleton-specific prompt          & 57.2     \\ 
		\midrule
		Ours                              & 58.2     \\
		\bottomrule
	\end{tabular}
	\caption{Ablation studies on 3D pose estimation module, semantic motion encoding and trainable skeleton-specific prompts. 
	}
	\label{tab_ablation}
\end{table}


\begin{table}
	\centering
	\setlength\tabcolsep{3pt}
		\resizebox{0.45\textwidth}{!}{
	\begin{tabular}{lccccc} 
		\toprule
		\multirow{2}{*}{\textbf{Method}} & \multirow{2}{*}{\textbf{Modality}} & \multicolumn{2}{c}{\textbf{x-sub}} & \multicolumn{2}{c}{\textbf{x-view}}  \\ 
		\cmidrule(lr){3-4} \cmidrule(lr){5-6}
		&                                    & 1\%           & 5\%                & 1\%           & 5\%                  \\ 
		\midrule
		ASSL~\cite{si2020adversarial}                                 & J                                  & -             & 57.3               & -             & 63.6                 \\
		ISC~\cite{thoker2021skeleton}                                 & J                                  & 35.7          & 59.6               & 38.1          & 65.7                 \\
		MCC~\cite{su2021self}                                         & J                                  & -             & 47.4               & -             & 53.3                 \\
		Hi-TRS~\cite{chen2022hierarchically}                          & J                                  & 39.1          & 63.3               & 42.9          & 68.3                 \\
		GL-Transformer~\cite{kim2022global}                           & J                                  & -             & 64.5               & -             & 68.5                 \\
		Colorization~\cite{yang2021skeleton}                          & J                                  & 48.3          & 65.7               & 52.5          & 70.3                 \\
		CrosSCLR~\cite{li20213d}                                      & J                                  & 48.6          & 67.7               & 49.8          & 70.6                 \\
		HiCo~\cite{dong2023hierarchical}                              & J                                  & 54.4          & -                  & 54.8          & -                    \\
		CPM~\cite{zhang2022contrastive}                               & J                                  & 56.7          & -                  & 57.5          & -                    \\
		CMD~\cite{mao2022cmd}                                         & J                                  & 50.6          & 71.0               & 53.0          & 75.3                 \\
		UmURL~\cite{sun2023unified}                                   & J\,+\,M\,+\,B                      & \textbf{59.6} & 74.6               & \textbf{60.3} & 78.6                 \\
		3s-AimCLR~\cite{guo2022contrastive}                           & J\,+\,M\,+\,B                      & 54.8          & -                  & 54.3          & -                    \\
		3s-CMD~\cite{mao2022cmd}                                      & J\,+\,M\,+\,B                      & 55.6          & 74.3               & 55.5          & 77.2                 \\
		\midrule
		Ours                                                          & J\,+\,C\,+ S                       & 55.0          & \textbf{76.3}      & 55.0          & \textbf{79.1}        \\
		\bottomrule
	\end{tabular}}
	\caption{Comparison with state-of-the-art methods on semi-supervised learning task on the NTU-60 dataset.}
	\label{tab_semi}
\end{table}

\subsection{Potential for Semi-Supervised Learning}
Following the experimental setup of the previous study~\cite{thoker2021skeleton}, we conduct experiments to evaluate the potential of the proposed method in semi-supervised learning. We first pre-train the encoder using our method, then randomly select 1\% and 5\% of the labeled training data during the fine-tuning phase to jointly optimize the encoder and classifier. To our surprise, the results vary greatly between the 1\% and 5\% settings. Our method outperforms all other methods in the 5\% setting, but only achieves relatively good results in the 1\% setting. After analysis, we believe that our method requires a certain amount of data for support. Too little data can lead to a decrease in model performance. Because the representation learning of heterogeneous skeletons is relatively complex and challenging for the model. However, as the amount of data increases, the improvement speed of our method is also very significant, allowing it to quickly surpass previous methods, which demonstrates the great potential of our method.

\subsection{Transferability for Action Recognition}
To validate that the representations learned through heterogeneous skeleton topologies offer improved generalization, we conduct transfer learning experiments. In these experiments, we first conduct pre-training on the model using unsupervised learning with the source dataset. Subsequently, we solely fine-tune the linear classification layer on the target dataset to assess its adaptability across various datasets.

\noindent\textbf{Transferability on 3D Skeleton Dataset:} 
As shown in Table~\ref{tab_transfer}, we use the NTU-60 and NTU-120 datasets as the source datasets and the PKU-MMD II dataset as the target dataset. Experiments are conducted under the x-sub setting. The results show that our method significantly outperforms other baseline approaches in transfer learning tasks. This advantage is mainly attributed to the superior representational capacity of the heterogeneous skeleton topology, which enables the learned features to transfer effectively and exhibit robust generalization capabilities.

\noindent\textbf{Transferability on 2D Skeleton Dataset:} 
We further validate the transferability of our method on heterogeneous skeletons by fine-tuning the classifier layer on the realistic skeleton dataset FineGYM~\cite{shao2020finegym}, which provides two-dimensional skeletons with 17-joint skeleton. As can be seen from Table~\ref{tab_finegym}, our method achieves competitive results on FineGYM in transfer learning task. This indicates that our method exhibits outstanding transferability and flexibility when dealing with heterogeneous skeleton. 

\begin{table}
	\centering
	\setlength\tabcolsep{2pt}
	\resizebox{0.45\textwidth}{!}{
	\begin{tabular}{lccc} 
		\toprule
		\multirow{2}{*}{\textbf{Method}} & \multirow{2}{*}{\textbf{Modality}} & \multicolumn{2}{c}{\textbf{Transfer to PKU-MMD II}}  \\ 
		\cmidrule(lr){3-4}
		&                                    & \textbf{NTU-60} & \textbf{NTU-120}                  \\ 
		\midrule
		LongT GAN~\cite{zheng2018unsupervised}                        & J                                  & 44.8            & -                                 \\
		M\^{2}L~\cite{lin2020ms2l}                                    & J                                  & 45.8            & -                                 \\
		ISC~\cite{thoker2021skeleton}                                 & J                                  & 45.9            & -                                 \\
		CrosSCLR~\cite{li20213d}                                      & J                                  & 54.0            & 52.8                              \\
		HiCo~\cite{dong2023hierarchical}                              & J                                  & 56.3            & 55.4                              \\
		CMD~\cite{mao2022cmd}                                         & J                                  & 56.0            & 57.0                              \\
		UmURL~\cite{sun2023unified}                                   & J\,+\,M\,+\,B                      & 59.7            & 58.5                              \\
		\midrule
		Ours                                                          & J\,+\,C\,+\,S                      & \textbf{64.3}   & \textbf{63.1}                     \\
		\bottomrule
	\end{tabular}}
	\caption{Comparison with state-of-the-art methods on the transfer learning for action recognition task.}
	\label{tab_transfer}
\end{table}

\begin{table}
	\centering
	\begin{tabular}{lcc} 
		\toprule
		\textbf{Method}  & \textbf{Modality}  & \textbf{Accuracy}   \\ 
		\midrule
		BEAR~\cite{deng2023large}           & RGB       & 69.6       \\
		MoLo~\cite{wang2023molo}            & RGB+Point & 73.3       \\
		SVT~\cite{ranasinghe2022self}       & RGB       & 62.3       \\
		Euclidean~\cite{han2019video}       & RGB       & 68.2       \\
		Hyperbolic~\cite{suris2021learning} & RGB       & 73.4       \\
		CARL~\cite{chen2022frame}           & RGB       & 41.8       \\
		\midrule
		Ours                                & Skeleton  & \textbf{75.3}  \\
		\bottomrule
	\end{tabular}
	\caption{Comparison of action recognition methods transferred from NTU-120 pretraining to the FineGYM dataset.}
	\label{tab_finegym}
\end{table}

\section{Conclusion}
\label{sec:conclusion}

In this work, we study data heterogeneity in human skeletons and introduce a unified framework for learning heterogeneous skeleton-based action representations. This framework consists of two components: heterogeneous skeleton processing and unified representation learning. We also design semantic motion coding to leverage the semantic information within skeletons. We take the three-dimensional 25-joint skeleton and the two-dimensional 17-joint skeleton as examples for self-supervised representation learning. Extensive experiments demonstrate the proposed model's ability to recognize human actions from different heterogeneous data. These heterogeneous skeletons are highly complementary for action recognition, and significantly enhance the performance through efficient early fusion. The proposed model also exhibits excellent transferability across various scenarios of skeleton-based action understanding. However, the shortcoming is that the model is limited to processing skeleton data for up to two people. 

\section*{Acknowledgments}
This work was supported by National Science Foundation of China (62172090, 62302093), Jiangsu Province Natural Science Fund (BK20230833), Start-up Research Fund of Southeast University (RF1028623097), and Big Data Computing Center of Southeast University.

{
    \small
    \bibliographystyle{ieeenat_fullname}
    \bibliography{main}

\begin{thebibliography}{52}
\providecommand{\natexlab}[1]{#1}
\providecommand{\url}[1]{\texttt{#1}}
\expandafter\ifx\csname urlstyle\endcsname\relax
  \providecommand{\doi}[1]{doi: #1}\else
  \providecommand{\doi}{doi: \begingroup \urlstyle{rm}\Url}\fi

\bibitem[Bardes et~al.(2021)Bardes, Ponce, and LeCun]{bardes2021vicreg}
Adrien Bardes, Jean Ponce, and Yann LeCun.
\newblock Vicreg: Variance-invariance-covariance regularization for
  self-supervised learning.
\newblock \emph{arXiv preprint arXiv:2105.04906}, 2021.

\bibitem[Chen et~al.(2022{\natexlab{a}})Chen, Wei, Li, and Cai]{chen2022frame}
Minghao Chen, Fangyun Wei, Chong Li, and Deng Cai.
\newblock Frame-wise action representations for long videos via sequence
  contrastive learning.
\newblock In \emph{Proceedings of the IEEE/CVF Conference on Computer Vision
  and Pattern Recognition}, pages 13801--13810, 2022{\natexlab{a}}.

\bibitem[Chen et~al.(2022{\natexlab{b}})Chen, Zhao, Yuan, Tian, Xia, Geng, Han,
  and Metaxas]{chen2022hierarchically}
Yuxiao Chen, Long Zhao, Jianbo Yuan, Yu Tian, Zhaoyang Xia, Shijie Geng, Ligong
  Han, and Dimitris~N Metaxas.
\newblock Hierarchically self-supervised transformer for human skeleton
  representation learning.
\newblock In \emph{European Conference on Computer Vision}, pages 185--202.
  Springer, 2022{\natexlab{b}}.

\bibitem[Chi et~al.(2022)Chi, Ha, Chi, Lee, Huang, and Ramani]{chi2022infogcn}
Hyung-gun Chi, Myoung~Hoon Ha, Seunggeun Chi, Sang~Wan Lee, Qixing Huang, and
  Karthik Ramani.
\newblock Infogcn: Representation learning for human skeleton-based action
  recognition.
\newblock In \emph{Proceedings of the IEEE/CVF Conference on Computer Vision
  and Pattern Recognition}, pages 20186--20196, 2022.

\bibitem[Deng et~al.(2023)Deng, Yang, and Chen]{deng2023large}
Andong Deng, Taojiannan Yang, and Chen Chen.
\newblock A large-scale study of spatiotemporal representation learning with a
  new benchmark on action recognition.
\newblock In \emph{Proceedings of the IEEE/CVF International Conference on
  Computer Vision}, pages 20519--20531, 2023.

\bibitem[Dong et~al.(2023)Dong, Sun, Liu, Chen, Liu, and
  Wang]{dong2023hierarchical}
Jianfeng Dong, Shengkai Sun, Zhonglin Liu, Shujie Chen, Baolong Liu, and Xun
  Wang.
\newblock Hierarchical contrast for unsupervised skeleton-based action
  representation learning.
\newblock In \emph{Proceedings of the AAAI Conference on Artificial
  Intelligence}, pages 525--533, 2023.

\bibitem[Du et~al.(2016)Du, Fu, and Wang]{du2016representation}
Yong Du, Yun Fu, and Liang Wang.
\newblock Representation learning of temporal dynamics for skeleton-based
  action recognition.
\newblock \emph{IEEE Transactions on Image Processing}, 25\penalty0
  (7):\penalty0 3010--3022, 2016.

\bibitem[Franco et~al.(2023)Franco, Mandica, Munjal, and
  Galasso]{francohyperbolic}
Luca Franco, Paolo Mandica, Bharti Munjal, and Fabio Galasso.
\newblock Hyperbolic self-paced learning for self-supervised skeleton-based
  action representations.
\newblock In \emph{International Conference on Learning Representations}, 2023.

\bibitem[Guo et~al.(2022)Guo, Liu, Chen, Liu, Wang, and
  Ding]{guo2022contrastive}
Tianyu Guo, Hong Liu, Zhan Chen, Mengyuan Liu, Tao Wang, and Runwei Ding.
\newblock Contrastive learning from extremely augmented skeleton sequences for
  self-supervised action recognition.
\newblock In \emph{Proceedings of the AAAI Conference on Artificial
  Intelligence}, pages 762--770, 2022.

\bibitem[Han et~al.(2019)Han, Xie, and Zisserman]{han2019video}
Tengda Han, Weidi Xie, and Andrew Zisserman.
\newblock Video representation learning by dense predictive coding.
\newblock In \emph{Proceedings of the IEEE/CVF International Conference on
  Computer Vision Workshops}, pages 0--0, 2019.

\bibitem[Kim et~al.(2022)Kim, Chang, Kim, and Choi]{kim2022global}
Boeun Kim, Hyung~Jin Chang, Jungho Kim, and Jin~Young Choi.
\newblock Global-local motion transformer for unsupervised skeleton-based
  action learning.
\newblock In \emph{European Conference on Computer Vision}, pages 209--225.
  Springer, 2022.

\bibitem[Li et~al.(2021)Li, Wang, Ni, Wang, Yang, and Zhang]{li20213d}
Linguo Li, Minsi Wang, Bingbing Ni, Hang Wang, Jiancheng Yang, and Wenjun
  Zhang.
\newblock 3d human action representation learning via cross-view consistency
  pursuit.
\newblock In \emph{Proceedings of the IEEE/CVF Conference on Computer Vision
  and Pattern Recognition}, pages 4741--4750, 2021.

\bibitem[Lin et~al.(2020)Lin, Song, Yang, and Liu]{lin2020ms2l}
Lilang Lin, Sijie Song, Wenhan Yang, and Jiaying Liu.
\newblock Ms2l: Multi-task self-supervised learning for skeleton based action
  recognition.
\newblock In \emph{Proceedings of the ACM international conference on
  multimedia}, pages 2490--2498, 2020.

\bibitem[Lin et~al.(2023)Lin, Zhang, and Liu]{lin2023actionlet}
Lilang Lin, Jiahang Zhang, and Jiaying Liu.
\newblock Actionlet-dependent contrastive learning for unsupervised
  skeleton-based action recognition.
\newblock In \emph{Proceedings of the IEEE/CVF Conference on Computer Vision
  and Pattern Recognition}, pages 2363--2372, 2023.

\bibitem[Lin et~al.(2025)Lin, Wu, Zhang, and Liu]{lin2025idempotent}
Lilang Lin, Lehong Wu, Jiahang Zhang, and Jiaying Liu.
\newblock Idempotent unsupervised representation learning for skeleton-based
  action recognition.
\newblock In \emph{European Conference on Computer Vision}, pages 75--92.
  Springer, 2025.

\bibitem[Liu et~al.(2019)Liu, Shahroudy, Perez, Wang, Duan, and
  Kot]{liu2019ntu}
Jun Liu, Amir Shahroudy, Mauricio Perez, Gang Wang, Ling-Yu Duan, and Alex~C
  Kot.
\newblock Ntu rgb+ d 120: A large-scale benchmark for 3d human activity
  understanding.
\newblock \emph{IEEE Transactions on Pattern Analysis and Machine
  Intelligence}, 42\penalty0 (10):\penalty0 2684--2701, 2019.

\bibitem[Liu et~al.(2020)Liu, Song, Liu, Li, and Hu]{liu2020benchmark}
Jiaying Liu, Sijie Song, Chunhui Liu, Yanghao Li, and Yueyu Hu.
\newblock A benchmark dataset and comparison study for multi-modal human action
  analytics.
\newblock \emph{ACM Transactions on Multimedia Computing, Communications, and
  Applications}, 16\penalty0 (2):\penalty0 1--24, 2020.

\bibitem[Mao et~al.(2022)Mao, Zhou, Lu, Deng, and Li]{mao2022cmd}
Yunyao Mao, Wengang Zhou, Zhenbo Lu, Jiajun Deng, and Houqiang Li.
\newblock Cmd: Self-supervised 3d action representation learning with
  cross-modal mutual distillation.
\newblock In \emph{European Conference on Computer Vision}, pages 734--752.
  Springer, 2022.

\bibitem[Mao et~al.(2023)Mao, Deng, Zhou, Fang, Ouyang, and Li]{mao2023masked}
Yunyao Mao, Jiajun Deng, Wengang Zhou, Yao Fang, Wanli Ouyang, and Houqiang Li.
\newblock Masked motion predictors are strong 3d action representation
  learners.
\newblock In \emph{Proceedings of the IEEE/CVF International Conference on
  Computer Vision}, pages 10181--10191, 2023.

\bibitem[Nie et~al.(2020)Nie, Liu, and Liu]{nie2020unsupervised}
Qiang Nie, Ziwei Liu, and Yunhui Liu.
\newblock Unsupervised 3d human pose representation with viewpoint and pose
  disentanglement.
\newblock In \emph{European Conference on Computer Vision}, pages 102--118.
  Springer, 2020.

\bibitem[Plizzari et~al.(2021)Plizzari, Cannici, and
  Matteucci]{plizzari2021skeleton}
Chiara Plizzari, Marco Cannici, and Matteo Matteucci.
\newblock Skeleton-based action recognition via spatial and temporal
  transformer networks.
\newblock \emph{Computer Vision and Image Understanding}, 208:\penalty0 103219,
  2021.

\bibitem[Radford et~al.(2021)Radford, Kim, Hallacy, Ramesh, Goh, Agarwal,
  Sastry, Askell, Mishkin, Clark, et~al.]{radford2021learning}
Alec Radford, Jong~Wook Kim, Chris Hallacy, Aditya Ramesh, Gabriel Goh,
  Sandhini Agarwal, Girish Sastry, Amanda Askell, Pamela Mishkin, Jack Clark,
  et~al.
\newblock Learning transferable visual models from natural language
  supervision.
\newblock In \emph{International Conference on Machine Learning}, pages
  8748--8763. PMLR, 2021.

\bibitem[Ranasinghe et~al.(2022)Ranasinghe, Naseer, Khan, Khan, and
  Ryoo]{ranasinghe2022self}
Kanchana Ranasinghe, Muzammal Naseer, Salman Khan, Fahad~Shahbaz Khan, and
  Michael~S Ryoo.
\newblock Self-supervised video transformer.
\newblock In \emph{Proceedings of the IEEE/CVF Conference on Computer Vision
  and Pattern Recognition}, pages 2874--2884, 2022.

\bibitem[Shah et~al.(2023)Shah, Roy, Shah, Mishra, Jacobs, Cherian, and
  Chellappa]{shah2023halp}
Anshul Shah, Aniket Roy, Ketul Shah, Shlok Mishra, David Jacobs, Anoop Cherian,
  and Rama Chellappa.
\newblock Halp: Hallucinating latent positives for skeleton-based
  self-supervised learning of actions.
\newblock In \emph{Proceedings of the IEEE/CVF Conference on Computer Vision
  and Pattern Recognition}, pages 18846--18856, 2023.

\bibitem[Shahroudy et~al.(2016)Shahroudy, Liu, Ng, and Wang]{shahroudy2016ntu}
Amir Shahroudy, Jun Liu, Tian-Tsong Ng, and Gang Wang.
\newblock Ntu rgb+ d: A large scale dataset for 3d human activity analysis.
\newblock In \emph{Proceedings of the IEEE Conference on Computer Vision and
  Pattern Recognition}, pages 1010--1019, 2016.

\bibitem[Shao et~al.(2020)Shao, Zhao, Dai, and Lin]{shao2020finegym}
Dian Shao, Yue Zhao, Bo Dai, and Dahua Lin.
\newblock Finegym: A hierarchical video dataset for fine-grained action
  understanding.
\newblock In \emph{Proceedings of the IEEE/CVF Conference on Computer Vision
  and Pattern Recognition}, pages 2616--2625, 2020.

\bibitem[Si et~al.(2020)Si, Nie, Wang, Wang, Tan, and Feng]{si2020adversarial}
Chenyang Si, Xuecheng Nie, Wei Wang, Liang Wang, Tieniu Tan, and Jiashi Feng.
\newblock Adversarial self-supervised learning for semi-supervised 3d action
  recognition.
\newblock In \emph{European Conference on Computer Visio}, pages 35--51.
  Springer, 2020.

\bibitem[Su et~al.(2020)Su, Liu, and Shlizerman]{su2020predict}
Kun Su, Xiulong Liu, and Eli Shlizerman.
\newblock Predict \& cluster: Unsupervised skeleton based action recognition.
\newblock In \emph{Proceedings of the IEEE/CVF Conference on Computer Vision
  and Pattern Recognition}, pages 9631--9640, 2020.

\bibitem[Su et~al.(2021)Su, Lin, and Wu]{su2021self}
Yukun Su, Guosheng Lin, and Qingyao Wu.
\newblock Self-supervised 3d skeleton action representation learning with
  motion consistency and continuity.
\newblock In \emph{Proceedings of the IEEE/CVF International Conference on
  Computer Vision}, pages 13328--13338, 2021.

\bibitem[Sun et~al.(2019)Sun, Xiao, Liu, and Wang]{sun2019deep}
Ke Sun, Bin Xiao, Dong Liu, and Jingdong Wang.
\newblock Deep high-resolution representation learning for human pose
  estimation.
\newblock In \emph{Proceedings of the IEEE/CVF Conference on Computer Vision
  and Pattern Recognition}, pages 5693--5703, 2019.

\bibitem[Sun et~al.(2023)Sun, Liu, Dong, Qu, Gao, Yang, Wang, and
  Wang]{sun2023unified}
Shengkai Sun, Daizong Liu, Jianfeng Dong, Xiaoye Qu, Junyu Gao, Xun Yang, Xun
  Wang, and Meng Wang.
\newblock Unified multi-modal unsupervised representation learning for
  skeleton-based action understanding.
\newblock In \emph{Proceedings of the ACM International Conference on
  Multimedia}, pages 2973--2984, 2023.

\bibitem[Sur{\'\i}s et~al.(2021)Sur{\'\i}s, Liu, and
  Vondrick]{suris2021learning}
D{\'\i}dac Sur{\'\i}s, Ruoshi Liu, and Carl Vondrick.
\newblock Learning the predictability of the future.
\newblock In \emph{Proceedings of the IEEE/CVF Conference on Computer Vision
  and Pattern Recognition}, pages 12607--12617, 2021.

\bibitem[Thoker et~al.(2021)Thoker, Doughty, and Snoek]{thoker2021skeleton}
Fida~Mohammad Thoker, Hazel Doughty, and Cees~GM Snoek.
\newblock Skeleton-contrastive 3d action representation learning.
\newblock In \emph{Proceedings of the ACM international conference on
  multimedia}, pages 1655--1663, 2021.

\bibitem[Wang and Wang(2017)]{wang2017modeling}
Hongsong Wang and Liang Wang.
\newblock Modeling temporal dynamics and spatial configurations of actions
  using two-stream recurrent neural networks.
\newblock In \emph{Proceedings of the IEEE Conference on Computer Vision and
  Pattern Recognition}, pages 499--508, 2017.

\bibitem[Wang and Wang(2018)]{wang2018beyond}
Hongsong Wang and Liang Wang.
\newblock Beyond joints: Learning representations from primitive geometries for
  skeleton-based action recognition and detection.
\newblock \emph{IEEE Transactions on Image Processing}, 27\penalty0
  (9):\penalty0 4382--4394, 2018.

\bibitem[Wang et~al.(2022)Wang, Wen, Si, Qian, and Wang]{wang2022contrast}
Peng Wang, Jun Wen, Chenyang Si, Yuntao Qian, and Liang Wang.
\newblock Contrast-reconstruction representation learning for self-supervised
  skeleton-based action recognition.
\newblock \emph{IEEE Transactions on Image Processing}, 31:\penalty0
  6224--6238, 2022.

\bibitem[Wang et~al.(2023)Wang, Zhang, Qing, Gao, Zhang, Zhao, and
  Sang]{wang2023molo}
Xiang Wang, Shiwei Zhang, Zhiwu Qing, Changxin Gao, Yingya Zhang, Deli Zhao,
  and Nong Sang.
\newblock Molo: Motion-augmented long-short contrastive learning for few-shot
  action recognition.
\newblock In \emph{Proceedings of the IEEE/CVF Conference on Computer Vision
  and Pattern Recognition}, pages 18011--18021, 2023.

\bibitem[Weng et~al.(2025)Weng, Wang, Wang, He, and Xie]{weng2024usdrl}
Wanjiang Weng, Hongsong Wang, Junbo Wang, Lei He, and Guosen Xie.
\newblock Usdrl: Unified skeleton-based dense representation learning with
  multi-grained feature decorrelation.
\newblock In \emph{Proceedings of the AAAI Conference on Artificial
  Intelligence}, 2025.

\bibitem[Wu et~al.(2024)Wu, Wu, Kittler, Xu, Ahmed, Awais, and Feng]{wu2024scd}
Cong Wu, Xiao-Jun Wu, Josef Kittler, Tianyang Xu, Sara Ahmed, Muhammad Awais,
  and Zhenhua Feng.
\newblock Scd-net: Spatiotemporal clues disentanglement network for
  self-supervised skeleton-based action recognition.
\newblock In \emph{Proceedings of the AAAI Conference on Artificial
  Intelligence}, pages 5949--5957, 2024.

\bibitem[Xiang et~al.(2023)Xiang, Li, Zhou, Wang, and
  Zhang]{xiang2023generative}
Wangmeng Xiang, Chao Li, Yuxuan Zhou, Biao Wang, and Lei Zhang.
\newblock Generative action description prompts for skeleton-based action
  recognition.
\newblock In \emph{Proceedings of the IEEE/CVF International Conference on
  Computer Vision}, pages 10276--10285, 2023.

\bibitem[Xu et~al.(2021)Xu, Shen, Wong, and Kankanhalli]{xu2021unsupervised}
Ziwei Xu, Xudong Shen, Yongkang Wong, and Mohan~S Kankanhalli.
\newblock Unsupervised motion representation learning with capsule
  autoencoders.
\newblock \emph{Advances in Neural Information Processing Systems},
  34:\penalty0 3205--3217, 2021.

\bibitem[Yan et~al.(2018)Yan, Xiong, and Lin]{yan2018spatial}
Sijie Yan, Yuanjun Xiong, and Dahua Lin.
\newblock Spatial temporal graph convolutional networks for skeleton-based
  action recognition.
\newblock In \emph{Proceedings of the AAAI Conference on Artificial
  Intelligence}, 2018.

\bibitem[Yang et~al.(2021)Yang, Liu, Lu, Er, and Kot]{yang2021skeleton}
Siyuan Yang, Jun Liu, Shijian Lu, Meng~Hwa Er, and Alex~C Kot.
\newblock Skeleton cloud colorization for unsupervised 3d action representation
  learning.
\newblock In \emph{Proceedings of the IEEE/CVF International Conference on
  Computer Vision}, pages 13423--13433, 2021.

\bibitem[Zhang et~al.(2022)Zhang, Hou, Zhang, and Li]{zhang2022contrastive}
Haoyuan Zhang, Yonghong Hou, Wenjing Zhang, and Wanqing Li.
\newblock Contrastive positive mining for unsupervised 3d action representation
  learning.
\newblock In \emph{European Conference on Computer Vision}, pages 36--51.
  Springer, 2022.

\bibitem[Zhang et~al.(2023{\natexlab{a}})Zhang, Lin, and
  Liu]{zhang2023hierarchical}
Jiahang Zhang, Lilang Lin, and Jiaying Liu.
\newblock Hierarchical consistent contrastive learning for skeleton-based
  action recognition with growing augmentations.
\newblock In \emph{Proceedings of the AAAI Conference on Artificial
  Intelligence}, pages 3427--3435, 2023{\natexlab{a}}.

\bibitem[Zhang et~al.(2023{\natexlab{b}})Zhang, Lin, and
  Liu]{zhang2023prompted}
Jiahang Zhang, Lilang Lin, and Jiaying Liu.
\newblock Prompted contrast with masked motion modeling: Towards versatile 3d
  action representation learning.
\newblock In \emph{Proceedings of the 31st ACM International Conference on
  Multimedia}, pages 7175--7183, 2023{\natexlab{b}}.

\bibitem[Zheng et~al.(2018)Zheng, Wen, Liu, Long, Dai, and
  Gong]{zheng2018unsupervised}
Nenggan Zheng, Jun Wen, Risheng Liu, Liangqu Long, Jianhua Dai, and Zhefeng
  Gong.
\newblock Unsupervised representation learning with long-term dynamics for
  skeleton based action recognition.
\newblock In \emph{Proceedings of the AAAI Conference on Artificial
  Intelligence}, 2018.

\bibitem[Zhou et~al.(2022)Zhou, Cheng, Li, Fang, Geng, Xie, and
  Keuper]{zhou2022hypergraph}
Yuxuan Zhou, Zhi-Qi Cheng, Chao Li, Yanwen Fang, Yifeng Geng, Xuansong Xie, and
  Margret Keuper.
\newblock Hypergraph transformer for skeleton-based action recognition.
\newblock \emph{arXiv preprint arXiv:2211.09590}, 2022.

\bibitem[Zhou et~al.(2023)Zhou, Duan, Rao, Su, and Wang]{zhou2023self}
Yujie Zhou, Haodong Duan, Anyi Rao, Bing Su, and Jiaqi Wang.
\newblock Self-supervised action representation learning from partial
  spatio-temporal skeleton sequences.
\newblock In \emph{Proceedings of the AAAI Conference on Artificial
  Intelligence}, pages 3825--3833, 2023.

\bibitem[Zhou et~al.(2024)Zhou, Yan, Cheng, Yan, Dai, and
  Hua]{zhou2024blockgcn}
Yuxuan Zhou, Xudong Yan, Zhi-Qi Cheng, Yan Yan, Qi Dai, and Xian-Sheng Hua.
\newblock Blockgcn: Redefine topology awareness for skeleton-based action
  recognition.
\newblock In \emph{Proceedings of the IEEE/CVF Conference on Computer Vision
  and Pattern Recognition}, pages 2049--2058, 2024.

\bibitem[Zhu et~al.(2023{\natexlab{a}})Zhu, Ma, Liu, Liu, Wu, and
  Wang]{zhu2023motionbert}
Wentao Zhu, Xiaoxuan Ma, Zhaoyang Liu, Libin Liu, Wayne Wu, and Yizhou Wang.
\newblock Motionbert: A unified perspective on learning human motion
  representations.
\newblock In \emph{Proceedings of the IEEE/CVF International Conference on
  Computer Vision}, pages 15085--15099, 2023{\natexlab{a}}.

\bibitem[Zhu et~al.(2023{\natexlab{b}})Zhu, Han, Yu, and Liu]{zhu2023modeling}
Yisheng Zhu, Hu Han, Zhengtao Yu, and Guangcan Liu.
\newblock Modeling the relative visual tempo for self-supervised skeleton-based
  action recognition.
\newblock In \emph{Proceedings of the IEEE/CVF International Conference on
  Computer Vision}, pages 13913--13922, 2023{\natexlab{b}}.

\end{thebibliography}
}


\end{document}